\documentclass[journal]{IEEEtran}
\usepackage[T1]{fontenc}
\usepackage{cite}
\usepackage{amsmath,amssymb,amsfonts}
\usepackage{booktabs,array,tabularx}
\usepackage{algorithm}
\usepackage[noend]{algpseudocode}
\usepackage[hidelinks]{hyperref}
\usepackage{tikz}
\usetikzlibrary{arrows.meta}          % for the -{Stealth[...]} arrows in Fig. rates
\usepackage{pgfplots}
\pgfplotsset{compat=1.18}

\providecolor{ink}{HTML}{1F2937}
\providecolor{accent}{HTML}{1F6FB2}
\providecommand{\FNz}{\mathcal{F}_{N}^{\emptyset}}   % false abstentions
\providecommand{\FNa}{\mathcal{F}_{N}^{\mathrm{ans}}}    % non-matching answers

\newlength{\omw}
\newlength{\omh}
\newcommand{\outcomematrix}[5]{%
  \begin{tikzpicture}[
      omcell/.style={draw=gray!60, line width=0.4pt, minimum width=\omw,
                     minimum height=\omh, inner sep=0pt, align=center,
                     font=\scriptsize},
      omok/.style={omcell, fill=accent!14},
      omerr/.style={omcell, fill=red!75!black!11}]
    \node[omok,  anchor=north west] at (0,0)         {{\tiny TP}\\#2};
    \node[omerr, anchor=north west] at (\omw,0)      {{\tiny FN}\\#3};
    \node[omerr, anchor=north west] at (0,-\omh)     {{\tiny FP}\\#4};
    \node[omok,  anchor=north west] at (\omw,-\omh)  {{\tiny TN}\\#5};
    \node[font=\scriptsize, anchor=south, inner sep=2pt,
          text height=1.6ex, text depth=0.4ex] at (\omw,0) {#1};
  \end{tikzpicture}%
}
\definecolor{ink}{HTML}{153B58}
\definecolor{pale}{HTML}{E9F0F5}
\definecolor{accent}{HTML}{1E7180}
\definecolor{warm}{HTML}{F5EBDD}
\newcommand{\TP}{\mathcal{T}_{P}}
\newcommand{\TN}{\mathcal{T}_{N}}
\newcommand{\FP}{\mathcal{F}_{P}}
\newcommand{\FN}{\mathcal{F}_{N}}
\newcommand{\CS}{\mathcal{C}_{S}}
\newcommand{\CR}{\mathcal{C}_{R}}
\newcommand{\EM}{\mathrm{EM}}
\newcommand{\SES}{\mathrm{SES}}
\newcommand{\RVS}{\mathrm{RVS}}
\newcommand{\ind}{\mathbf{1}}
\newcommand{\NA}{\texttt{[]}}
\newcommand{\Judge}{\textsc{Judge}}
\newcommand{\grp}[1]{\textsf{#1}}
\newcommand{\doi}[1]{\href{https://doi.org/#1}{doi: \nolinkurl{#1}}}
\newcommand{\webref}[2]{\href{#1}{#2}}

\hypersetup{
    pdftitle={SAGE-QA: Semantic-Aware, Grounded Evaluation for Closed-Context Question Answering on SQuAD 2.0},
    pdfauthor={Chandra N Sekharan, Vishakan Umapathy, and Abhishek Phadke}
}

\title{SAGE-QA: Semantic-Aware, Grounded \\Evaluation for Closed-Context Question \\Answering on SQuAD 2.0}
\author{Chandra N Sekharan, Vishakan Umapathy, and Abhishek Phadke%
\thanks{C. N. Sekharan and V. Umapathy are with the Department of Computer Science, Texas A\&M University--Corpus Christi, Corpus Christi, TX, USA.}%
\thanks{A. Phadke is with the School of Engineering and Computing, Christopher Newport University, Newport News, VA, USA.}}

\begin{document}
\maketitle
\begin{abstract}
Closed-context question answering (QA) involves a system to generate an answer strictly supported by a supplied passage or to abstain when the passage lacks adequate evidence. The Stanford Question Answering Dataset (SQuAD 2.0) is a key benchmark widely used to evaluate machine reading comprehension under these constraints~\cite{squad2,squad1,leaderboard}. Traditionally, model performance on SQuAD 2.0 is measured via Exact Match (EM), which requires character-level agreement with human reference spans after lexical normalization. For generative large language models (LLMs), however, an EM failure does not necessarily indicate an erroneous response; it may instead reflect a valid paraphrase, an alternative passage-supported answer, or a  hallucination. Prior results on the official leader board achieved a maximum EM of approximately 91\%~\cite{leaderboard} but evaluating  generative models requires separating lexical divergence from factual error. Our primary contribution of the paper is the introduction of \textbf{SAGE-QA} (\textbf{S}emantic-\textbf{A}ware, \textbf{G}rounded \textbf{E}valuation for \textbf{Q}uestion \textbf{A}nswering), a multi-tiered evaluation framework that combines reference-based Semantic Equivalence Scoring (SES) and passage-based Re-Validation Scoring (RVS). 
Across a fixed set of 10,000 SQuAD 2.0 validation examples, we compute baseline EM and decision-matrix F1 directly from True Positive (TP), True Negative (TN), False Positive (FP), and False Negative (FN) outcome counts across zero-shot, closed-weight frontier models and open-weight models (untuned zero-shot and fine-tuned versions). 
 In zero-shot evaluation, closed-weight frontier models Claude Opus 4.6 and Gemini 3.1 Pro achieve 85.21 and 86.66 EM, outperforming untuned open-weight models Llama 3 8B (56.88 EM), gpt-oss-20b (66.28 EM), and Llama 3.3 70B (71.39 EM). However, parameter-efficient fine-tuning yields gains of 21.19 to 34.06 EM points over untuned baselines, elevating fine-tuned Llama 3 8B, gpt-oss-20b, and Llama 3.3 70B to 90.94, 87.47, and 94.24 EM scores, respectively. Additionally, when we audit the zero-shot frontier models using SAGE-QA with an external LLM judge, the  dataset-wide SES scores (on a 0--100 scale) are 92.81 for Claude Opus 4.6 and 91.67 for Gemini 3.1 Pro, while the corresponding hybrid dataset-wide RVS scores are 97.96 and 98.02. This combines benchmark-based credit with passage-based judge assessments. Our findings demonstrate that fine-tuning open-weight models perform the best to surpass previous best known EM benchmarks. 
\end{abstract}


\section{Introduction}
\IEEEPARstart{C}{losed}-context question answering (QA) tasks a language model with answering a question using only the evidence provided in a companion passage, or returning an explicit abstention token (\NA) when no answer is supported. We explore this question because  QA systems needs to satisfy the information need expressed by the query while following the strict evidence boundary set by the source text. The Stanford Question Answering Dataset (SQuAD)~\cite{squad2,squad1} serves as the benchmark for this work, comprising comprehension questions authored by crowd workers over Wikipedia passages. SQuAD 2.0 extends the original 100,000 answerable questions of SQuAD 1.1 with over 50,000 adversarial, unanswerable questions designed to closely resemble answerable queries~\cite{squad2}. SQuAD 2.0  requires a solution that not only extracts or generates accurate answers when sufficient evidence is present, but also reliably abstains when the passage does not support a conclusion. For clarity, reference answers are to be considered as ground-truths because they are provided by human crowd workers. 

While traditional AI retrieval models select contiguous character spans from the passage, modern generative models frequently formulate complete phrases or valid paraphrases. Under standard Exact Match (EM) evaluation, any deviation from the exact wording or span boundaries chosen by crowd annotators is penalized as a failure. To examine disagreements between exact reference matching and LLM-based assessment, we propose the \textbf{SAGE-QA} evaluation framework, which extends standard confusion-matrix metrics to include two complementary LLM-as-a-judge diagnostics triggered whenever a prediction fails EM.
\begin{enumerate}
    \item \textbf{Semantic Equivalence Scoring (SES):} SES uses evaluation on answerable questions where the model's answer  fails EM. SES employs an external judge LLM to assess whether the generated answer preserves the semantic meaning of at least one (ground-truth) reference answer.
    \item \textbf{Re-Validation Scoring (RVS):} This evaluates on all  predictions that fail EM (spanning both answerable and benchmark-unanswerable questions), a judge LLM is used to assess  whether the candidate answer directly addresses the question and if it is fully grounded in the source passage. The judge receives no reference answers (including not answerable or [] type of answers). But scoring uses all of the reference answers to check against the model answer. 
\end{enumerate}
To avoid direct self-evaluation that may be biased, we execute both SES and RVS using an external evaluator model (\texttt{gpt-6-luna}~\cite{luna}) that does not participate as an answer generator in our benchmark. 

Architecturally, we maintain a strict separation between answer generation and answer evaluation. On the generation side, we investigate automated prompt optimization 
for zero-shot, closed-weight frontier models (Claude Opus 4.6 and Gemini 3.1 Pro) and evaluate open-weight models (Llama 3 8B, Llama 3.3 70B, and gpt-oss-20b) in both 
untuned, zero-shot and parameter-efficient fine-tuned (LoRA and QLoRA) configurations. This allows us to compare closed- and open-weight models zero-shot, while isolating the gain of fine-tuning. On the evaluation side, we retain strict lexical matching (EM and outcome-count F1) and add separate judge-based assessments of reference-meaning agreement (SES) and passage support (RVS). Our diagnostic contribution is to characterize which non-matching responses receive judge-assigned credit, where the two rubrics disagree, and how the evaluation criterion affects aggregate scores and model comparisons. 

\section{Background and Related Work}
\subsection{Benchmark Evolution and Extractive Baselines}

\noindent
SQuAD 1.1 focused exclusively on span extraction for answerable queries~\cite{squad1} but SQuAD 2.0 introduced unanswerable questions to penalize models that rely on superficial keyword matching or heuristic guessing~\cite{squad2}. Early neural architectures, such as BiDAF~\cite{bidaf} and discriminative pre-trained encoders based on BERT~\cite{bert} and RoBERTa~\cite{roberta}, framed closed-context QA as a token-classification problem over passage start and end indices. Table~\ref{tab:historical} summarizes key reference points from the official SQuAD 2.0 leaderboard~\cite{leaderboard}, including human baseline performance and IE-Net expressed as a percentage.  Note that the leaderboard's $\mathrm{F1}_{\mathrm{reported}}$ metric measures token-level bag-of-words overlap; we provide it here solely for historical context, as it is mathematically distinct from the decision-matrix F1 defined in Section~\ref{sec:framework}.

\begin{table}[!t]
\centering\footnotesize
\caption{Historical Leaderboard Context on the Official Hidden Test Set~\cite{leaderboard}.}
\label{tab:historical}
\begin{tabular}{lrr}\toprule
\textbf{System} & \textbf{EM} & $\mathbf{F1_{\mathrm{reported}}}$\\\midrule
Human Performance & 86.831 & 89.452\\
IE-Net (Discriminative Ensemble) & 90.939 & 93.214\\\bottomrule
\end{tabular}
\end{table}

When generative LLMs are evaluated on span-retrieval benchmarks, the relationship between token overlap and answer correctness breaks down~\cite{kamalloo,bulian,adlakha}. Consider a representative example from the SQuAD 2.0 repository~\cite{hfdataset}.

\begin{quote}
\small

\textbf{Context:} \emph{``To further highlight the difference between a problem and an instance, consider the following instance of the decision version of the traveling salesman problem: Is there a route of at most 2000 kilometres passing through all of Germany's 15 largest cities? The quantitative answer to this particular problem instance is of little use for solving other instances of the problem, such as asking for a round trip through all sites in Milan whose total length is at most 10 km. For this reason, complexity theory addresses computational problems and not particular problem instances.''}\par
\textbf{Question:} \emph{What is one example of an instance that the quantitative answer to the traveling salesman problem fails to answer?}\par
{\raggedright\textbf{Human Reference Answers:} \texttt{["round trip through all sites in Milan", "asking for a round trip through all sites in Milan whose total length is at most 10 km", "a round trip through all sites in Milan whose total length is at most 10 km"]}\par}
\end{quote}
A generative model that responds with \emph{``finding a round trip of at most 10 km visiting every site in Milan''} provides a factually exact, passage-grounded answer that nonetheless receives a score of zero under Exact Match. Notice that the three human references differ only in whether they include the leading words  (\emph{``asking for a...''}) or the trailing constraint (\emph{``whose total length is at most 10 km''}); consequently, any valid generative paraphrase or alternative span choice is penalized as a complete failure under surface-level lexical matching.

\subsection{Semantic Equivalence and Passage Grounding}

\noindent
Kamalloo et al.~\cite{kamalloo} demonstrated that lexical matching severely underestimates generative LLMs in open-domain QA because models frequently produce valid paraphrases or varied granularity spans. However, their analysis focused on always-answerable open-domain queries; SAGE-QA examines evaluation disagreement in closed-context QA with adversarial unanswerable questions, separating judge-assessed reference equivalence ($\SES$ on $\FNa$) from judge-assessed passage support ($\RVS$ on $\FP \cup \FNa$), as defined  precisely later. Embedding-based metrics such as BERTScore~\cite{bertscore} compute contextual token similarity, but  recent frameworks such as G-Eval~\cite{geval} and MT-Bench~\cite{mtbench} leverage LLMs directly as automated judges. Studies of annotation artifacts and label errors motivate closer examination of benchmark failures~\cite{northcutt,gururangan}; they do not establish that disagreements observed under our judging protocol are annotation errors.

Building toward semantic evaluation in QA, Bulian et al.~\cite{bulian} formulated question-aware answer-equivalence judgments and trained the learned BEM (BERT Matching) evaluator. Their asymmetric formulation can accept a candidate answer that expands upon or improves a reference, but our SES metric enforces a symmetric, scaled reference-meaning rubric anchored by strict mutual equivalence, in a logical sense. Complementing reference equivalence, Adlakha et al.~\cite{adlakha} demonstrated the importance of separating reference correctness from context faithfulness in instruction-following QA. The RVS judge applies a passage-based rubric to assess this faithfulness criterion within a closed-context setting, without receiving reference answers. 

Directly relevant to our benchmark dataset, Bahak et al.~\cite{bahak} evaluated ChatGPT (\texttt{gpt-3.5-turbo}) as a closed-context QA system on SQuAD 1.1 and SQuAD 2.0 using standard character-normalized EM, Recall, and token-overlap F1. They found that ChatGPT trailed task-specific discriminative models by a wide margin, achieving only 44.4\% EM (67\% F1) on SQuAD 1.1 and plummeting to 18.12\% EM (56\% F1) on SQuAD 2.0. Bahak et al.\ attributed this collapse on SQuAD 2.0 to the model generating substantive responses on unanswerable questions, automatically classifying all such False Positives as model hallucinations. 

Our study examines three related aspects: first, programmatic prompt optimization and structured abstention schemas yield zero-shot,  frontier EM above 85\%--86\%; second, the RVS judge accepts over 90\% of benchmark False Positives in our frontier cohort, documenting disagreement with benchmark answerability labels,  and third, parameter-efficient, fine-tuning enables open-weight generative models to surpass the 90.58\%--90.94\% EM ceiling from prior literature.

\subsection{Prompt Optimization, Adaptation, and Evaluator Bias}
To obtain optimal performance from generative models, recent work has shifted from manual prompt engineering to programmatic prompt compilation using frameworks such as DSPy~\cite{dspy}, MIPROv2~\cite{mipro}, GEPA~\cite{gepa}, and automated migration optimizers~\cite{optimizer}. In parallel, parameter-efficient fine-tuning techniques, notably LoRA~\cite{lora} and QLoRA~\cite{qlora}, enable open-weight models to adapt rapidly to domain-specific output formatting and abstention calibration.

When auditing model outputs with LLM judges, however, evaluators can exhibit self-preference bias by favoring generations produced by their own model family, and sensitivity to prompt structure and candidate ordering~\cite{selfbias,faireval}. We avoid direct self-evaluation by employing a single, external judge model (\texttt{gpt-6-luna}~\cite{luna}) that generates none of the candidate answers in our study, and enforce a uniform, blind input schema across all evaluated systems. 

\section{Definitions and The SAGE-QA Evaluation Framework}
\label{sec:framework}

We first define a partitioning of $N$ QA evaluation items into four mutually exclusive outcome sets which will be used throughout the rest of the paper.  
\begin{itemize}
    \item \textbf{True Positive ($\TP$):} The question is answerable ($a_i=1$), the model does not abstain ($z_i=0$), and the normalized prediction matches a reference ($e_i=1$).
    \item \textbf{True Negative ($\TN$):} The question is unanswerable ($a_i=0$) and the model abstains ($z_i=1$).
    \item \textbf{False Positive ($\FP$):} The question is unanswerable ($a_i=0$), but the model supplies an answer ($z_i=0$).
    \item \textbf{False Negative ($\FN$):} The question is answerable ($a_i=1$), but the model either abstains ($z_i=1$) or supplies an answer that fails EM ($z_i=0$ and $e_i=0$).
\end{itemize}

To provide an end-to-end diagnostic of closed-context QA evaluation, we structure SAGE-QA around three sequential evaluation stages. First, we interpret each raw answer field and retain an abstention flag before applying lexical normalization for Exact Match. We then use the flag and the EM result to partition the $N=10{,}000$ benchmark examples into correctness-aware lexical outcome categories ($\TP, \TN, \FP$, and $\FN$, where $\FN$ is further split into false abstentions $\FNz$ and answered non-matches $\FNa$). Second, we route eligible lexical failures concurrently to two separate LLM-as-a-judge diagnostics: Semantic Equivalence Scoring ($\CS = \FNa$) and Re-Validation Scoring ($\CR = \FP \cup \FNa$). These are consistent with the fact that the SES algorithm assigns a score of 0 to $\FP$ and false abstentions, while RVS assigns a score of 0 to false abstentions but uses a scaled(gradually graded) approach for $\FP$ and $\FNa$. Third, we aggregate both eligible case-level means and dataset-wide scores that credit $\TP$ and $\TN$ automatically while avoiding any double-counting on the overlapping set $\FNa$. These hybrid aggregates combine benchmark-matching credit with judge-assigned rubric scores. Table~\ref{tab:pipeline_flow} summarizes the complete routing architecture.

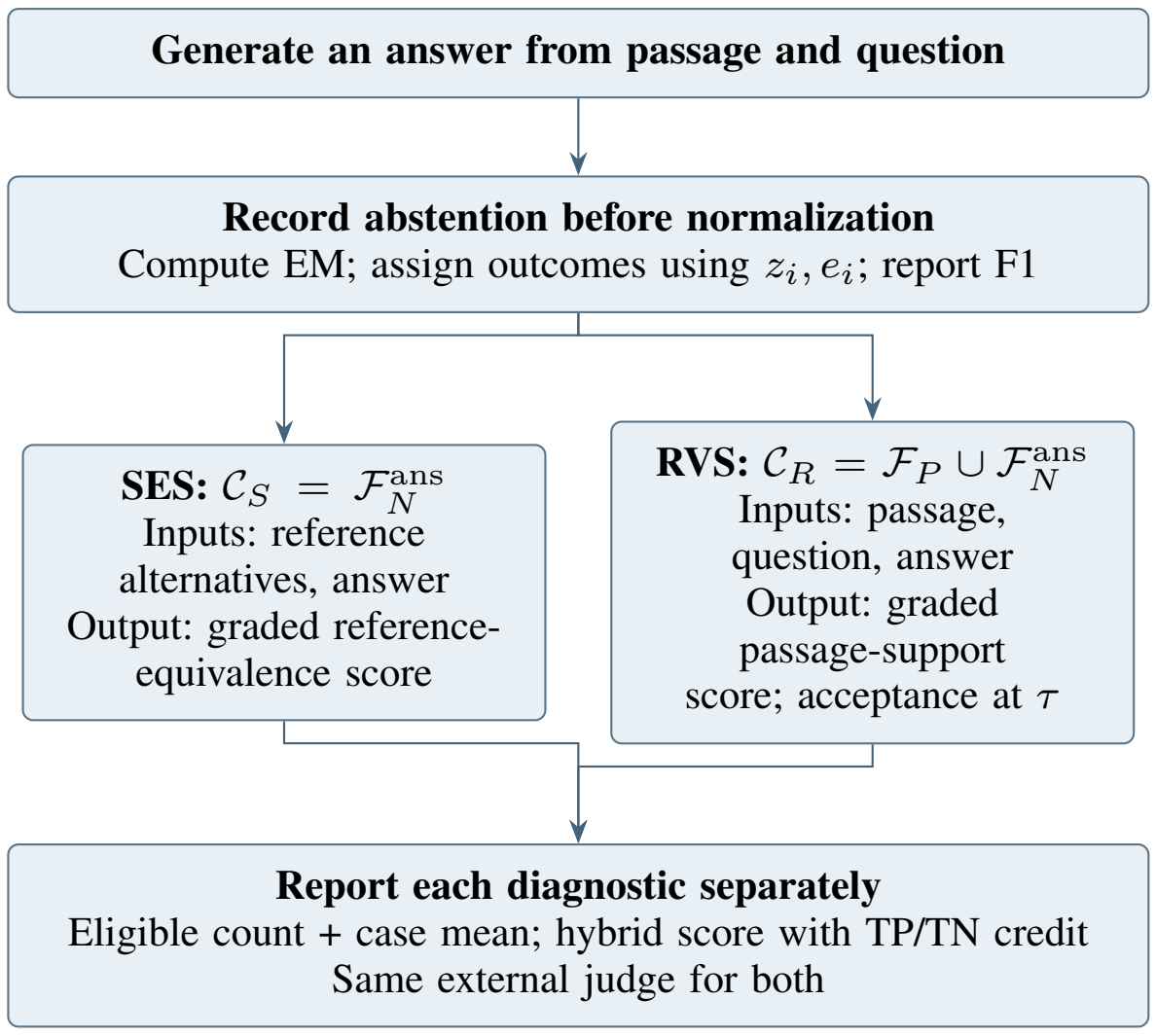
\begin{figure}[!t]
\centering
\begin{tikzpicture}[font=\footnotesize,
box/.style={draw=ink!70,rounded corners=2pt,fill=pale,align=center,text width=7.4cm,inner sep=5pt},
smallbox/.style={box,text width=3.2cm,minimum height=1.8cm},arr/.style={-{Stealth[length=1.6mm]},ink!80}]
\node[box] (gen) at (0,0) {\textbf{Generate an answer from passage and question}};
\node[box] (lex) at (0,-1.3) {\textbf{Record abstention before normalization}\\Compute EM; assign outcomes using $z_i,e_i$; report F1};
\node[smallbox] (ses) at (-2,-3.6) {\textbf{SES: $\CS=\FNa$}\\Inputs: reference alternatives, answer\\Output: graded reference-equivalence score};
\node[smallbox] (rvs) at (2,-3.6) {\textbf{RVS: $\CR=\FP\cup\FNa$}\\Inputs: passage, question, answer\\Output: graded passage-support score; acceptance at $\tau$};
\node[box] (agg) at (0,-6.0) {\textbf{Report each diagnostic separately}\\Eligible count + case mean; hybrid score with TP/TN credit\\Same external judge for both };
\draw[arr] (gen)--(lex);
\draw[arr] (lex.south)--++(0,-.15)-|(ses.north);
\draw[arr] (lex.south)--++(0,-.15)-|(rvs.north);
\draw[arr] (ses.south)--++(0,-.15)-|(agg.north);
\draw[arr] (rvs.south)--++(0,-.15)-|(agg.north);
\end{tikzpicture}
\caption{SAGE-QA pipeline.  SES and RVS overlap on answered FN cases; RVS is not conditioned on SES failure. TP/TN retain benchmark-based full credit without judging. False abstentions receive no judge assessment and zero credit in both  diagnostic scores.}
\label{fig:pipeline}
\end{figure}

\begin{table*}[!t]
\centering\footnotesize
\caption{Overview of the SAGE-QA Diagnostic Routing Pipeline.}
\label{tab:pipeline_flow}
\begin{tabularx}{\textwidth}{l l X}
\toprule
\textbf{Stage} & \textbf{Target Pool / Metric} & \textbf{Operational Definition and Inputs} \\
\midrule
1. Generation & All $N=10{,}000$ items & Model generates a raw answer field from $(c_i, q_i)$, using \NA\ to indicate abstention. \\
2. Interpretation and Lexical Match & $\EM, \mathrm{F1}$, $\TP, \TN, \FP, \FN$ & Determine $z_i$ from the raw answer and retain interpreted text $\hat{y}_i$. Normalize only for EM; assign outcomes using $a_i,z_i,e_i$, and split $\FN$ using $z_i$. \\
3a. SES Judge & $\CS = \FNa$ & External judge scores semantic equivalence $s_i \in [0,1]$ using only $(G_i, \hat{y}_i)$. \\
3b. RVS Judge & $\CR = \FP \cup \FNa$ & Same external judge scores passage support $r_i \in [0,1]$ (and binary acceptance at $\tau=0.95$) using only $(c_i, q_i, \hat{y}_i)$. \\
4. Aggregation & $\SES_{\mathrm{all}}, \RVS_{\mathrm{all}}$ & Report eligible case means and hybrid dataset-wide scores combining benchmark-based credit for $\TP$ and $\TN$ with judge-assigned credit, without double-counting overlapping $\FNa$ cases. \\
\bottomrule
\end{tabularx}
\end{table*}

\subsection{Prediction Format and Exact Matching}
For each evaluation example $i \in \{1, \dots, N\}$, let $c_i$ denote the context passage, $q_i$ the question, $G_i$ the set of human reference answers, and $a_i \in \{0, 1\}$ the ground-truth answerability label. Each evaluated model supplies a raw answer (response). Before any lexical normalization, we set the abstention flag $z_i=1$ if this response is an empty list, the string \NA, an empty string, or a missing value; otherwise, we set $z_i=0$. The scoring scripts retain the complementary flag \texttt{pred\_is\_answerable}, corresponding to $1-z_i$. We define $\hat{y}_i$ as the interpreted prediction text: the empty string when $z_i=1$, and the extracted answer text otherwise. The flag $z_i$ is retained separately and is never recomputed from normalized text.

The two diagnostics use separate requests to the same judge: SES receives only $(G_i, \hat{y}_i)$, while RVS receives only $(c_i, q_i, \hat{y}_i)$. The prediction text supplied to the judge has not undergone SQuAD lexical normalization. Neither request includes the benchmark answerability flag $a_i$.

We apply standard SQuAD lexical normalization~\cite{hfmetric} by stripping articles, punctuation, and excess whitespace while lowercasing for only the Exact Match comparison of the interpreted prediction and reference answers. This operation does not alter the retained $z_i$ or $\hat{y}_i$. In particular, a non-abstaining answer that becomes empty after normalization is not reclassified as an abstention. Let $n(\cdot)$ denote this normalization function and let $\widetilde{G}_i$ denote the set of normalized references for example $i$ (where $\widetilde{G}_i = \{\text{``''}\}$ for unanswerable questions). Exact Match is then defined using the indicator function $\ind\{\cdot\}$ maximized over $\widetilde{G}_i$:
\begin{equation}
e_i = \max_{g \in \widetilde{G}_i} \ind\{n(\hat{y}_i) = g\}, \qquad
\EM = \frac{100}{N} \sum_{i=1}^{N} e_i.
\label{eq:em}
\end{equation}

\subsection{ Count-Based F1}

Since $\FN$ incorporates both failure to answer and failure to match a reference lexically, we partition $\FN$ into a disjoint union:
\begin{equation}
\FN = \FNz \mathbin{\dot\cup} \FNa,
\label{eq:fnsplit}
\end{equation}
where $\FNz = \{i : a_i=1,\ z_i=1\}$ contains false abstentions on answerable questions and $\FNa = \{i : a_i=1,\ z_i=0,\ e_i=0\}$ contains answered non-matches. On our 10,000-example evaluation split, these categories satisfy the constraints:
\begin{align}
|\TP| + |\FN| &= 4{,}990, \quad |\TN| + |\FP| = 5{,}010, \label{eq:margins}\\
\EM &= 100 \frac{|\TP| + |\TN|}{N}. \label{eq:emcounts}
\end{align}

From these correctness-aware cardinalities, we compute decision-matrix precision ($P$), recall ($R$), and percentage F1 ($\mathrm{F1}$).
\begin{equation}
P = \frac{|\TP|}{|\TP| + |\FP|}, \qquad
R = \frac{|\TP|}{|\TP| + |\FN|},
\label{eq:pr}
\end{equation}
\begin{equation}
\mathrm{F1} = 100 \frac{2PR}{P+R} = 100 \frac{2|\TP|}{2|\TP| + |\FP| + |\FN|}.
\label{eq:f1}
\end{equation}
Unlike traditional SQuAD token-overlap $\mathrm{F1}_{\mathrm{reported}}$, which awards partial credit for sub-span lexical intersection on individual questions, our decision-matrix $\mathrm{F1}$ in Eq.~\eqref{eq:f1} evaluates corpus-level classification performance directly from the cardinalities $|\TP|$, $|\FP|$, and $|\FN|$. Under this formulation, both false abstentions ($\FNz$) and non-matching answers ($\FNa$) penalize recall equally through their combined cardinality $|\FN|$, whereas True Negatives ($\TN$) contribute to $\EM$ in Eq.~\eqref{eq:emcounts} but are excluded from $\mathrm{F1}$. Consequently, when a model achieves higher accuracy on unanswerable questions ($|\TN|$) than on answerable ones ($|\TP|$), its decision-matrix $\mathrm{F1}$ can be slightly lower than its $\EM$---a structural property that we observe empirically across several evaluated models.

To characterize error trade-offs conditioned on ground-truth answerability, we also define the false positive rate (FP rate) and false negative rate (FN rate) over their respective 5,010 and 4,990 denominators:
\begin{equation}
\begin{aligned}
\mathrm{FP\ rate} &= 100 \frac{|\FP|}{|\FP| + |\TN|},\\
\mathrm{FN\ rate} &= 100 \frac{|\FN|}{|\TP| + |\FN|}.
\end{aligned}
\label{eq:rates}
\end{equation}

\subsection{SES: Reference-Meaning Agreement}
Semantic Equivalence Scoring (SES) asks the judge to assess whether an answered EM failure on an answerable question preserves the semantic meaning of at least one human reference. Consequently, we define its eligible evaluation pool strictly as the set of substantive non-matching predictions on answerable questions:
\begin{equation}
\CS = \FNa.
\label{eq:seselig}
\end{equation}
For each example $i \in \CS$, the external judge model $\theta$ receives only the human reference alternatives and the extracted answer, computing a graded equivalence score:
\begin{equation}
s_i = E^{\mathrm{SES}}_{\theta}(G_i, \hat{y}_i) \in [0, 1].
\label{eq:sesjudge}
\end{equation}
We treat $s_i \in [0, 1]$ as score using a structured rubric as indicated below.  Because the members of $G_i$ represent alternative valid annotations rather than fragments that must be jointly reproduced, the judge evaluates $\hat{y}_i$ against each reference $g \in G_i$ independently and retains the maximum score: $s_i = \max_{g \in G_i} \Judge(g, \hat{y}_i)$. We enforce a five-tier graded rubric:
\begin{itemize}
\item $1.00$: Mutual semantic equivalence without material omission or addition.
\item $0.95\text{--}0.99$: Strong equivalence with only minor, nonessential phrasing differences.
\item $0.80\text{--}0.94$: Partial equivalence exhibiting substantial overlap alongside a meaningful omission or addition.
\item $0.20\text{--}0.79$: Weak agreement with major semantic divergences despite topically related content.
\item $0.00$: Unrelated or contradictory meaning.
\end{itemize}
Algorithm~\ref{alg:ses} formalizes the end-to-end SES evaluation procedure across all $N$ benchmark instances.

\begin{algorithm*}[!t]
\small
\caption{Semantic Equivalence Scoring (SES) Procedure}
\label{alg:ses}
\begin{algorithmic}[1]
\Require For each example $i \in \{1, \dots, N\}$: lexical outcome label $o_i \in \{\grp{TP}, \grp{TN}, \grp{FP}, \grp{FN}\}$, reference set $G_i$, abstention flag $z_i$ determined from the raw answer field before normalization, and interpreted prediction text $\hat{y}_i$.
\Ensure Eligible case mean $\SES_{\mathrm{cases}}$ and dataset-wide score $\SES_{\mathrm{all}}$.
\For{each example $i = 1, \dots, N$}
    \If{$o_i \in \{\grp{TP}, \grp{TN}\}$}
        \State $s_i \gets 1$ \Comment{Automatic full credit for exact matches}
    \ElsIf{$o_i = \grp{FP}$ \textbf{or} $z_i = 1$}
        \State $s_i \gets 0$ \Comment{Zero credit for $\FP$ and false abstentions $\FNz$}
    \Else \Comment{Eligible answered EM failure ($i \in \CS = \FNa$)}
        \State $s_i \gets \max_{g \in G_i} \Judge(g, \hat{y}_i)$ \Comment{Graded reference-meaning equivalence in $[0,1]$}
        \State Resubmit request if $s_i$ is missing or outside $[0,1]$ to enforce schema validity
    \EndIf
\EndFor
\State \Return $\SES_{\mathrm{cases}} = \frac{100}{|\CS|}\sum_{i \in \CS} s_i$ and $\SES_{\mathrm{all}} = \frac{100}{N}\sum_{i=1}^{N} s_i$
\end{algorithmic}
\end{algorithm*}

Letting $\bar{s} = \frac{1}{|\CS|}\sum_{i \in \CS} s_i$ denote the unscaled mean over the eligible pool $\CS$, we express the case-level mean and the chained dataset-wide score, both on a 0--100 scale, as:
\begin{align}
\SES_{\mathrm{cases}} &= \frac{100}{|\CS|} \sum_{i \in \CS} s_i = 100\,\bar{s}, \label{eq:sescases}\\
\SES_{\mathrm{all}} &= \frac{100}{N} \Bigl(|\TP| + |\TN| + \sum_{i \in \CS} s_i\Bigr) \nonumber\\
&= \EM + 100\frac{|\CS|}{N}\bar{s} = \EM + \frac{|\CS|}{N} \SES_{\mathrm{cases}}. \label{eq:sesall}
\end{align}
As shown in Eq.~\eqref{eq:sesall}, exact matches ($\TP$ and $\TN$) receive full credit ($1.0$), eligible non-matching answers ($\FNa$) receive $s_i$, and false positives ($\FP$) as well as false abstentions ($\FNz$) receive $0$.

The graded SES scores and the graded RVS scores defined below express mean assigned credit on a 0--100 scale. Multiplication by 100 rescales that credit; it does not convert partial credit into a count of correct answers. We therefore report graded results as scores, rather than percentages of correct answers, and differences between graded scores as score points. Percentages are retained for proportions of examples, including thresholded judge acceptance rates.

\subsection{RVS: Passage-Based Assessment}
Re-Validation Scoring (RVS) asks the judge to assess whether a substantive model answer directly addresses the question and is fully supported by the passage, without consulting reference answers or benchmark answerability labels. We define its eligible evaluation pool as the union of all substantive predictions that fail EM:
\begin{equation}
\CR = \FP \cup \FNa.
\label{eq:rvselig}
\end{equation}
For each example $i \in \CR$, the RVS judge receives exclusively the passage, question, and candidate answer, producing a passage-support score:
\begin{equation}
r_i = E^{\mathrm{RVS}}_{\theta}(c_i, q_i, \hat{y}_i) \in [0, 1].
\label{eq:rvsjudge}
\end{equation}
The judge is instructed to first assess from $c_i$ alone whether $q_i$ is answerable and what evidence $c_i$ provides; it then evaluates $\hat{y}_i$ jointly for \emph{groundedness} (every claim is stated or clearly implied by $c_i$, without external knowledge or contradiction) and \emph{completeness} (the response directly and fully answers $q_i$). RVS assigns scores using the same numerical bands as SES ($1.00$ fully grounded and complete; $0.95\text{--}0.99$ strong; $0.80\text{--}0.94$ partial; $0.20\text{--}0.79$ weak; $0.00$ unsupported, contradicted, or off-question). For binary acceptance analysis, we pre-declare a stringent threshold $\tau = 0.95$ and define $r_i^{\tau} = \ind\{r_i \geq \tau\}$. Algorithm~\ref{alg:rvs} summarizes the complete RVS evaluation workflow.

\begin{algorithm*}[!t]
\small
\caption{Re-Validation Scoring (RVS) Procedure}
\label{alg:rvs}
\begin{algorithmic}[1]
\Require For each example $i \in \{1, \dots, N\}$: lexical outcome label $o_i$, passage $c_i$, question $q_i$, abstention flag $z_i$ determined from the raw answer field before normalization, and interpreted prediction text $\hat{y}_i$.
\Ensure Judge mean on eligible cases $\RVS_{\mathrm{cases}}$ and hybrid dataset-wide score $\RVS_{\mathrm{all}}$.
\For{each example $i = 1, \dots, N$}
    \If{$o_i \in \{\grp{TP}, \grp{TN}\}$}
        \State $r_i \gets 1$ \Comment{Automatic full credit for exact matches}
    \ElsIf{$z_i = 1$} \Comment{False abstention on answerable question ($i \in \FNz$)}
        \State $r_i \gets 0$ \Comment{Zero credit for false abstentions $\FNz$; no judge call}
    \Else \Comment{Eligible substantive EM failure ($i \in \CR = \FP \cup \FNa$)}
        \State $r_i \gets \Judge(c_i, q_i, \hat{y}_i)$ \Comment{Graded passage groundedness and completeness in $[0,1]$}
        \State Resubmit request if $r_i$ is missing or outside $[0,1]$ to enforce schema validity
    \EndIf
\EndFor
\State \Return $\RVS_{\mathrm{cases}} = \frac{100}{|\CR|}\sum_{i \in \CR} r_i$ and $\RVS_{\mathrm{all}} = \frac{100}{N}\sum_{i=1}^{N} r_i$
\end{algorithmic}
\end{algorithm*}

Letting $\bar{r} = \frac{1}{|\CR|}\sum_{i \in \CR} r_i$ denote the unscaled mean over the eligible pool $\CR$, we write the judge mean on eligible cases and the hybrid dataset-wide score, both on a 0--100 scale, as:
\begin{align}
\RVS_{\mathrm{cases}} &= \frac{100}{|\CR|} \sum_{i \in \CR} r_i = 100\,\bar{r}, \label{eq:rvscases}\\
\RVS_{\mathrm{all}} &= \frac{100}{N} \Bigl(|\TP| + |\TN| + \sum_{i \in \CR} r_i\Bigr) \nonumber\\
&= \EM + 100\frac{|\CR|}{N}\bar{r} = \EM + \frac{|\CR|}{N} \RVS_{\mathrm{cases}}, \label{eq:rvsall}
\end{align}
where the binary acceptance counterparts replace $r_i$ with $r_i^{\tau}$. False abstentions ($\FNz$) lie outside $\CR$ and receive zero credit in both the graded and binary dataset-wide RVS scores.

The binary aggregate $\RVS^{\tau}_{\mathrm{all}}$ is the percentage of all $N$ examples credited by the combined benchmark-and-judge rules: benchmark $\TP$ and $\TN$ cases receive credit automatically, and eligible cases receive credit when $r_i\geq\tau$. It is not an independently verified percentage of correct answers. The eligible-pool acceptance rate, in contrast, uses $|\CR|$ as its denominator and counts only thresholded judge decisions.

Throughout the RVS analysis, ``accepted'' means that a judge score meets the specified threshold $\tau=0.95$. Scores below that threshold are described as ``below threshold,'' not automatically as ``unsupported.'' Because the rubric evaluates both passage support and completeness, a response can fall below the threshold for reasons other than lack of support. The lowest score band ($r_i<0.20$) is reported separately from all below-threshold responses and does not establish a verified hallucination category.

Here, \emph{reference-free} describes only the inputs to individual RVS judge calls for $i \in \CR$. Benchmark reference answers and answerability labels determine the eligible pool $\CR$ and the outcome categories used for automatic credit. Consequently, $\RVS_{\mathrm{cases}}$ is a judge mean conditional on a benchmark-selected pool, and $\RVS_{\mathrm{all}}$ is a hybrid dataset-wide score combining benchmark-based credit for $\TP$ and $\TN$ with judge-assigned credit on eligible cases. The $\TP$ and $\TN$ cases receive full credit without separate assessment by the RVS judge. This benchmark dependence also applies to the binary aggregate $\RVS^{\tau}_{\mathrm{all}}$; excluding references from judge inputs does not make case selection or dataset-wide aggregation reference-free.  

\section{Experimental Setup}
\subsection{Dataset Partition and Model Cohorts}
We conduct our primary evaluation on a fixed subset of $N = 10{,}000$ examples from the public SQuAD 2.0 validation dataset ~\cite{hfdataset}, comprising 4,990 answerable and 5,010 unanswerable questions. This held-out evaluation set is shared identically across all zero-shot and fine-tuned models. For our fine-tuning experiments, we train on a separate, balanced subset of 87,000 examples drawn exclusively from the official SQuAD 2.0 training data, ensuring zero data leakage into the 10,000-example evaluation set.

We evaluate three distinct model cohorts:
\begin{itemize}
    \item \textbf{Zero-Shot Frontier Group:} Claude Opus 4.6~\cite{claude} and Gemini 3.1 Pro~\cite{gemini}. For this cohort, we retain full per-example prediction logs, FN subgroup partitions ($\FNz$ vs.\ $\FNa$), and individual judge scores ($s_i, r_i$).
    \item \textbf{Zero-Shot Open-Weight Group (Untuned Baselines):} Meta-Llama-3-8B-Instruct~\cite{llama8}, Llama-3.3-70B-Instruct~\cite{llama70}, 
	and gpt-oss-20b~\cite{gptoss}. Evaluating these models out-of-the-box establishes a direct zero-shot comparison against the closed-weight frontier models and
	provides a matched untuned baseline for our fine-tuning experiments, retaining full outcome counts and FN subgroup splits ($\FNz$ vs.\ $\FNa$) under reference-based 
	scoring.
    \item \textbf{Fine-Tuned Open-Weight Group:} Llama 3 8B~\cite{llama8}, Llama 3.3 70B~\cite{llama70}, and gpt-oss-20b~\cite{gptoss}. Because reference-based confusion-matrix counts suffice to quantify exact lexical adaptation gains over our matched untuned baselines, our analysis of the fine-tuned cohort focuses on reference-based metrics ($\EM, \mathrm{F1}$, and category error rates), whereas per-example SES and RVS judge audits are restricted to the zero-shot frontier group.
\end{itemize}

\subsection{Prompt Optimization and Zero-Shot Execution Protocol}
Prior to full-scale benchmarking, we conducted a preliminary prompt-optimization study comparing three human-authored prompts, an OpenAI-optimized prompt~\cite{optimizer}, and prompts compiled programmatically via DSPy MIPROv2~\cite{dspy,mipro} and GEPA~\cite{gepa}; because the DSPy GEPA optimizer achieved the highest pilot performance (69.60 EM and 80.70 $\mathrm{F1}_{\mathrm{reported}}$ against SQuAD references), we standardized on this GEPA-compiled prompt across all subsequent zero-shot evaluations.

In our main zero-shot benchmark, Claude Opus 4.6 and Gemini 3.1 Pro process all 10,000 passage-and-question pairs via their respective provider Batch APIs~\cite{claudebatch,geminibatch}. Similarly, the untuned open-weight models---Llama 3 8B Instruct (via Hugging Face/Featherless-AI), Llama 3.3 70B Instruct 
(via Hugging Face/OVHCloud), and gpt-oss-20b (via Groq)---are evaluated zero-shot across the exact same 10,000 examples. 
Each model is constrained via structured-output schemas to return a strict two-field JSON object containing the example identifier and the extracted answer (or \NA). All 10,000 requests per model across all five zero-shot runs succeeded and parsed cleanly. 

As an implementation-consistency check, we examined all 20,000 answers from the Claude Opus 4.6 and Gemini 3.1 Pro. Both scoring scripts determine abstention from the raw answer before applying normalization for EM, and the check found no misclassified abstentions or outcome assignments. 

\subsection{Fine-Tuning Protocol and Judge Execution}
We adapted the three open-weight models for one epoch over the 87,000-example training split using assistant-response cross-entropy supervision and the 8-bit AdamW optimizer. Llama 3 8B was fine-tuned via LoRA with Unsloth~\cite{lora,unsloth}; gpt-oss-20b used 4-bit quantization with LoRA; and Llama 3.3 70B employed QLoRA with 4-bit NF4 quantization, BF16 compute, rank-stabilized LoRA~\cite{qlora}, cosine learning-rate scheduling, a warmup ratio of $0.03$, and weight decay of $0.01$. Table~\ref{tab:hyper} summarizes the hyperparameter configurations.

\begin{table}[!t]
\centering\footnotesize
\caption{Hyperparameter Configurations for Fine-Tuned Open-Weight Models (All Trained for 1 Epoch).}
\label{tab:hyper}
\begin{tabular}{lrrrrr}\toprule
 & \textbf{LoRA} & & \textbf{Learning} & \textbf{Batch} & \textbf{Max Seq.}\\
\textbf{Model} & \textbf{Rank ($r$)} & $\boldsymbol{\alpha}$ & \textbf{Rate} & \textbf{Size} & \textbf{Len.}\\\midrule
Llama 3 8B & 128 & 32 & $5 \times 10^{-5}$ & 32 & 512\\
gpt-oss-20b & 128 & 32 & $3 \times 10^{-5}$ & 64 & 512\\
Llama 3.3 70B & 32 & 64 & $1 \times 10^{-5}$ & $60^{*}$ & 1,024\\\bottomrule
\end{tabular}
\par\vspace{2pt}{\footnotesize $^{*}$Effective batch size via gradient accumulation.}
\end{table}

For the SAGE-QA diagnostic phase on the zero-shot frontier cohort, we executed \texttt{gpt-6-luna}~\cite{luna} at medium reasoning effort via the OpenAI Batch API~\cite{openaibatch}, enforcing a strict two-field JSON response (identifier and numerical score in $[0,1]$). Across both models, the eligible diagnostic pools comprised $|\CS| = 888$ and $|\CR| = 1{,}314$ cases for Claude Opus 4.6, and $|\CS| = 599$ and $|\CR| = 1{,}163$ cases for Gemini 3.1 Pro. Together, these SES and RVS analyses comprise 3,964 judge API requests, all of which returned valid in-range scores.

\section{Experimental Results and Diagnostic Analysis}

\subsection{Baseline Reference-Based Performance}
\noindent
Table~\ref{tab:master_baseline} summarizes the reference-based outcomes for all eight configurations on the same 10,000 examples. We highlight three findings; the table provides the complete counts, error rates, and scores.

\begin{table*}[!t]
\centering\footnotesize
\caption{Unified Baseline Results and Category Error Rates on 10,000 SQuAD 2.0 Validation Examples.}
\label{tab:master_baseline}
\begin{tabular}{llrrrrrrrr}\toprule
\textbf{Cohort} & \textbf{Model} & \textbf{TP} & \textbf{FN ($\FNz / \FNa$)} & \textbf{FP} & \textbf{TN} & \textbf{FP Rate (\%)} & \textbf{FN Rate (\%)} & \textbf{EM} & \textbf{F1}\\\midrule
Zero-Shot (Frontier) & Claude Opus 4.6 & 3,937 & 1,053 (165 / 888) & 426 & 4,584 & 8.50 & 21.10 & 85.21 & 84.19\\
Zero-Shot (Frontier) & Gemini 3.1 Pro & 4,220 & 770 (171 / 599) & 564 & 4,446 & 11.26 & 15.43 & \textbf{86.66} & \textbf{86.35}\\\midrule
Zero-Shot (Open-Weight) & Llama 3 8B (Untuned) & 3,236 & 1,754 (107 / 1,647) & 2,558 & 2,452 & 51.06 & 35.15 & 56.88 & 60.01\\
Zero-Shot (Open-Weight) & Llama 3.3 70B (Untuned) & 3,576 & 1,414 (120 / 1,294) & 1,447 & 3,563 & \textbf{28.88} & 28.34 & \textbf{71.39} & \textbf{71.43}\\
Zero-Shot (Open-Weight) & gpt-oss-20b (Untuned) & 3,610 & 1,380 (112 / 1,268) & 1,992 & 3,018 & 39.76 & \textbf{27.66} & 66.28 & 68.16\\\midrule
Fine-Tuned (Open-Weight) & Llama 3 8B (Fine-Tuned) & 4,681 & 309 (---) & 597 & 4,413 & 11.92 & 6.19 & 90.94 & 91.18\\
Fine-Tuned (Open-Weight) & Llama 3.3 70B (Fine-Tuned) & 4,684 & 306 (---) & 270 & 4,740 & \textbf{5.39} & \textbf{6.13} & \textbf{94.24} & \textbf{94.21}\\
Fine-Tuned (Open-Weight) & gpt-oss-20b (Fine-Tuned) & 4,039 & 951 (---) & 302 & 4,708 & 6.03 & 19.06 & 87.47 & 86.57\\\bottomrule
\end{tabular}
\end{table*}

\noindent\textbf{Gemini's higher EM comes with more answers on unanswerable questions.}
Relative to Claude, Gemini produces 283 additional exact matches on answerable questions but also answers 138 more questions designated unanswerable by SQuAD. The net advantage is 145 exact matches, or 1.45 EM percentage points (Fig.~\ref{fig:zs}). False abstentions are similar between the models; most of the difference on answerable questions lies in substantive answers that fail lexical matching. Their EM comparison therefore reflects a trade-off between reference matching and abstention.

\begin{figure}[!t]
\centering
\outcomematrix{Claude Opus 4.6 (ZS)}{3,937}{1,053}{426}{4,584}
\hspace{10mm}
\outcomematrix{Gemini 3.1 Pro (ZS)}{4,220}{770}{564}{4,446}
\caption{Zero-shot frontier outcome matrices. The upper row contains answerable examples (TP, FN); the lower row contains unanswerable examples (FP, TN). FN comprises 
false abstentions and non-matching answers (Claude $165 + 888$; Gemini $171 + 599$).}
\label{fig:zs}
\end{figure}

\noindent\textbf{Untuned, open-weight models frequently substantively answer when the benchmark expects abstention.}
Their FP rates range from 28.88\% to 51.06\%, compared with 8.50\%--11.26\% for the frontier models. On answerable questions, their failures predominantly involve nonmatching answers rather than abstentions (Fig.~\ref{fig:zs_open}). These two patterns account for their lower EM, but lexical counts alone cannot distinguish paraphrases from substantive answer errors. We have not applied SES/RVS to this cohort, so the semantic composition of these failures remains unresolved.

\begin{figure}[!t]
\centering
\outcomematrix{Llama 3 8B (ZS)}{3,236}{1,754}{2,558}{2,452}
\hspace{1mm}
\outcomematrix{Llama 3.3 70B (ZS)}{3,576}{1,414}{1,447}{3,563}
\hspace{1mm}
\outcomematrix{gpt-oss-20b (ZS)}{3,610}{1,380}{1,992}{3,018}
\caption{Zero-shot untuned open-weight outcome matrices. Without fine-tuning, open-weight models exhibit high FP counts on unanswerable questions and high non-matching $\FNa$ counts ($1{,}647$, $1{,}294$, and $1{,}268$) alongside low false abstentions ($\FNz$ of $107$, $120$, and $112$).}
\label{fig:zs_open}
\end{figure}

\noindent\textbf{Fine-tuning improves both categories of benchmark outcomes.}
Across the three open-weight models, adaptation increases EM by 21.19--34.06 percentage points and reduces both FP and FN counts (Fig.~\ref{fig:ft}). The two fine-tuned Llama models illustrate the importance of abstention: they achieve almost identical TP counts, yet the 70B model correctly abstains on 327 more unanswerable questions. This accounts for nearly all of its EM advantage over the 8B model. These comparisons establish improved reference-based outcomes on our evaluation subset.

\begin{figure}[!t]
\centering
\outcomematrix{Llama 3 8B (FT)}{4,681}{309}{597}{4,413}
\hspace{1mm}
\outcomematrix{Llama 3.3 70B (FT)}{4,684}{306}{270}{4,740}
\hspace{1mm}
\outcomematrix{gpt-oss-20b (FT)}{4,039}{951}{302}{4,708}
\caption{Fine-tuned open-weight outcome matrices, with the same row and category definitions as Figs.~\ref{fig:zs} and~\ref{fig:zs_open}. Every upper row sums to 
4,990 and every lower row to 5,010.}
\label{fig:ft}
\end{figure}

\subsection{SAGE-QA Diagnostic Results: SES and RVS}
\label{sec:judged}
We apply the judge-based diagnostics to the two zero-shot frontier models. Table~\ref{tab:judged} reports case means and dataset-wide scores using Eqs.~\eqref{eq:sesall} and~\eqref{eq:rvsall}; Table~\ref{tab:bands} shows the underlying score distributions.

\begin{table}[!t]
\centering\footnotesize
\caption{Judge-Based SES and RVS Diagnostic Summaries for the Zero-Shot Frontier Cohort ($\tau = 0.95$).}
\label{tab:judged}
\begin{tabularx}{\columnwidth}{@{}>{\raggedright\arraybackslash}Xrr@{}}\toprule
 & \textbf{Claude} & \textbf{Gemini}\\
\textbf{Diagnostic Metric} & \textbf{Opus 4.6} & \textbf{3.1 Pro}\\\midrule
Baseline EM (Table~\ref{tab:master_baseline}) & 85.21 & 86.66\\\midrule
Eligible SES Pool $|\CS| = |\FNa|$ & 888 & 599\\
$\SES_{\mathrm{cases}}$ (Mean, 0--100) & 85.56 & 83.68\\
$\SES_{\mathrm{all}}$ (Graded Dataset-Wide, 0--100) & \textbf{92.81} & 91.67\\\midrule
Eligible RVS Pool $|\CR| = |\FP \cup \FNa|$ & 1,314 & 1,163\\
$\RVS_{\mathrm{cases}}$ (Judge Mean on Eligible Cases, 0--100) & 97.01 & 97.66\\
$\RVS_{\mathrm{all}}$ (Hybrid Dataset-Wide, Graded, 0--100) & 97.96 & \textbf{98.02}\\
RVS Binary Acceptance Rate at $\tau=0.95$ (\%) & 93.68 & 92.86\\
$\RVS^{\tau}_{\mathrm{all}}$ (Hybrid Dataset-Wide, Binary Acceptance, \%) & \textbf{97.52} & 97.46\\\bottomrule
\end{tabularx}
\end{table}

\noindent\textbf{Reference-equivalence scoring reverses the EM ranking.}
Although Gemini leads under EM, Claude receives the higher dataset-wide SES score: 92.81 versus 91.67. Claude has both a larger eligible pool and a higher mean SES score within that pool. Reference-equivalence credit therefore changes the comparison, demonstrating that the lexical ranking is not preserved under this grading criterion.

\noindent\textbf{Passage-based scoring produces very similar aggregate scores.}
The graded dataset-wide RVS scores are 97.96 for Claude and 98.02 for Gemini, a gap of 0.06 score points. Binary acceptance at $\tau=0.95$ slightly reverses that ordering. These hybrid scores combine automatic benchmark credit with judge-assigned credit; their proximity does not establish equal answer correctness. Rather, the comparison depends on the evaluation criterion and treatment of partial credit.

\begin{table}[!t]
\centering\footnotesize
\setlength{\tabcolsep}{4pt}
\caption{Distribution of Judge Score Bands Across Eligible Diagnostic Pools.}
\label{tab:bands}
\begin{tabular}{@{}lrrrrr@{}}\toprule
 & & $\mathbf{0.95\text{--}}$ & $\mathbf{0.80\text{--}}$ & $\mathbf{0.20\text{--}}$ & \\
\textbf{Evaluated Pool ($n$)} & $\mathbf{1.00}$ & $\mathbf{0.99}$ & $\mathbf{0.94}$ & $\mathbf{0.79}$ & $\mathbf{<0.20}$\\\midrule
\multicolumn{6}{l}{\textit{Claude Opus 4.6}}\\
SES on $\CS = \FNa$ ($888$) & 96 & 219 & 493 & 50 & 30\\
RVS on $\FP$ ($426$) & 373 & 12 & 18 & 3 & 20\\
RVS on $\FNa$ ($888$) & 825 & 21 & 30 & 6 & 6\\\midrule
\multicolumn{6}{l}{\textit{Gemini 3.1 Pro}}\\
SES on $\CS = \FNa$ ($599$) & 87 & 160 & 275 & 52 & 25\\
RVS on $\FP$ ($564$) & 496 & 21 & 31 & 4 & 12\\
RVS on $\FNa$ ($599$) & 546 & 17 & 30 & 5 & 1\\\bottomrule
\end{tabular}
\end{table}

\noindent\textbf{The score distributions explain the difference between SES and RVS.}
On answered FN cases, partial equivalence is the most frequent SES category, whereas full credit is the most frequent RVS category (Table~\ref{tab:bands}). At $\tau=0.95$, SES accepts 35.5\% of Claude's and 41.2\% of Gemini's answered nonmatches; RVS accepts 95.3\% and 94.0\%, respectively. RVS also accepts over 90\% of each model's answers to questions designated unanswerable by SQuAD. These findings document judge disagreement with lexical matching and reference annotations, without independently resolving it. Scores below threshold can reflect incomplete answers as well as lack of passage support, so they should not all be described as unsupported.

\subsection{Paired SES/RVS Decision Comparison}
The distributions above describe each diagnostic separately. Table~\ref{tab:paired} compares their decisions on the same answered FN responses. RVS alone accepts 536 Claude responses and 321 Gemini responses, whereas SES alone accepts only five for each model. Thus, the main disagreement consists of responses receiving high passage-support credit without high reference-equivalence credit. Both assessments come from the same judge under different inputs and rubrics; this pattern identifies a difference between the criteria, not independent confirmation of answer correctness.

\begin{table}[!t]
\centering\footnotesize
\caption{Paired Judge Decisions on $\FNa$ at $\tau=0.95$.}
\label{tab:paired}
\begin{tabularx}{\columnwidth}{@{}>{\raggedright\arraybackslash}Xrr@{}}\toprule
 & \textbf{Claude} & \textbf{Gemini}\\
\textbf{Diagnostic Category} & \textbf{Opus 4.6} & \textbf{3.1 Pro}\\\midrule
\multicolumn{3}{l}{\textit{Paired Decisions on Answered FN Cases ($\FNa$)}}\\
Accepted by Both ($\SES \geq 0.95 \land \RVS \geq 0.95$) & 310 & 242\\
Accepted by SES Only ($\SES \geq 0.95 \land \RVS < 0.95$) & 5 & 5\\
Accepted by RVS Only ($\SES < 0.95 \land \RVS \geq 0.95$) & 536 & 321\\
Accepted by Neither ($\SES < 0.95 \land \RVS < 0.95$) & 37 & 31\\
\bottomrule
\end{tabularx}
\end{table}

\section{Discussion and Limitations}
\subsection{Evaluation Disagreement Beyond Lexical Matching}

\noindent\textbf{Lexical failure does not identify the type of failure.}
An EM failure establishes that a response does not match the reference annotations, but does not explain why. By separating reference equivalence from passage support, SES and RVS distinguish how the judge assesses these responses under two different criteria. Their paired decisions (Table~\ref{tab:paired}) reveal distinctions that a lexical score alone cannot express.

\noindent\textbf{Model comparisons depend on what we measure.}
The ranking changes in Table~\ref{tab:judged} show that the evaluation criterion affects the comparison. Dataset-wide diagnostic scores also retain automatic credit for benchmark successes. Consequently, a higher score or a smaller gap under SES or RVS does not, by itself, establish greater answer correctness or equivalence between models.

\noindent\textbf{Disagreements identify cases for examination.}
We use these diagnostics to locate responses for closer review while preserving the original benchmark outcomes. Their practical value is to make disagreements visible and organize subsequent examination.

\subsection{Limitations of our Research}
Our semantic-equivalence and passage-support assessments rely on one external LLM judge (\texttt{gpt-6-luna}) and therefore we interpret SES and RVS as judge-based diagnostic scores rather than independently verified measures of answer correctness. Responses accepted despite failing benchmark criteria identify cases for further examination, but do not by themselves establish annotation errors or model hallucination rates. Using a judge that generated none of the evaluated responses avoids direct self-evaluation ~\cite{selfbias,faireval}. Moreover, while our untuned-versus-fine-tuned comparisons isolate the exact lexical gains of parameter-efficient adaptation across all three open-weight models, we executed per-example SES and RVS assessments only on the zero-shot closed-weight frontier cohort. Extending these diagnostics to the untuned open-weight models (particularly their 1,268--1,647 $\FNa$ cases and 1,447--2,558 $\FP$ cases), together with independent human adjudication, would help examine the contributions of formatting divergence and comprehension errors to their lexical deficit that we defer to future studies as mentioned below.

\section{Conclusion}
In this paper, we presented SAGE-QA, a diagnostic evaluation framework for closed-context question answering that combines strict lexical matching ($\EM$ and 
decision-matrix $\mathrm{F1}$) with judge-assessed reference-meaning agreement (SES) and passage-based assessment (RVS). Evaluating eight configurations across 
10,000 SQuAD 2.0 validation examples, we demonstrated two complementary findings. First, under strict lexical evaluation, zero-shot closed-weight frontier 
models (Gemini 3.1 Pro at 86.66 EM and Claude Opus 4.6 at 85.21 EM) outperform untuned open-weight models (56.88 to 71.39 EM) by 15 to 30 EM points 
due to severe untuned over-answering on unanswerable queries; however, single-epoch parameter-efficient fine-tuning yields gains of +21.19 to +34.06 EM points, making the fine-tuned models excel in performance over prior results. Second, when we audit zero-shot frontier 
 EM failures with an external LLM judge under SAGE-QA, the graded dataset-wide SES scores (on a 0--100 scale) are 92.81 (Claude Opus 4.6) and 91.67 (Gemini 3.1 Pro), and the corresponding hybrid dataset-wide RVS scores are 
 97.96 and 98.02. These RVS aggregates combine benchmark-based full credit for $\TP$ and $\TN$ with passage-based judge scores on eligible cases. The graded RVS gap is 0.06 score points under this protocol, and the judge accepts over 90\% of each model's benchmark false positives. These results characterize how evaluation criteria affect scores, rankings, and assessments of non-matching responses while preserving the benchmark labels. 
 In future work, we plan to calibrate SES and RVS against human adjudication across multiple judge ensembles 
and extend full per-example diagnostic auditing to both untuned and fine-tuned open-weight architectures.

\end{document}